\documentclass[]{cas-sc}

\usepackage[numbers]{natbib}
\usepackage{url}            
\usepackage{booktabs}       
\usepackage{amsfonts}       
\usepackage{nicefrac}       
\usepackage{microtype}      
\usepackage{amsmath}
\usepackage{graphicx}       
\usepackage{subcaption}
\usepackage{multirow}
\usepackage{caption}
\usepackage{array}

\begin{document}
\let\WriteBookmarks\relax
\def\floatpagepagefraction{1}
\def\textpagefraction{.001}
\shorttitle{Wang and Shalaby}
\shortauthors{Wang and Shalaby}

\title [mode = title]{Transit Pulse: Utilizing Social Media as a Source for Customer Feedback and Information Extraction with Large Language Model}

\author[1]{Jiahao Wang}
\ead{jhope.wang@mail.utoronto.ca}
\author[1]{Amer Shalaby}
\ead{amer.shalaby@utoronto.ca}
\address[1]{Transit Analytics Lab (TAL), Department of Civil and Mineral Engineering (Transportation Engineering, University of Toronto, 35 St. George Street, Toronto, Ontario M5S 1A4 Canada}

\begin{abstract}
Users of the transit system flood social networks daily with messages that contain valuable insights crucial for improving service quality. These posts help transit agencies quickly identify emerging issues. Parsing topics and sentiments is key to gaining comprehensive insights to foster service excellence. However, the volume of messages makes manual analysis impractical, and standard NLP techniques like Term Frequency-Inverse Document Frequency (TF-IDF) fall short in nuanced interpretation. Traditional sentiment analysis separates topics and sentiments before integrating them, often missing the interaction between them. This incremental approach complicates classification and reduces analytical productivity.
To address these challenges, we propose a novel approach to extracting and analyzing transit-related information, including sentiment and sarcasm detection, identification of unusual system problems, and location data from social media. Our method employs Large Language Models (LLM), specifically Llama 3, for a streamlined analysis free from pre-established topic labels. To enhance the model's domain-specific knowledge, we utilize Retrieval-Augmented Generation (RAG), integrating external knowledge sources into the information extraction pipeline.
We validated our method through extensive experiments comparing its performance with traditional NLP approaches on user tweet data from the real world transit system. Our results demonstrate the potential of LLMs to transform social media data analysis in the public transit domain, providing actionable insights and enhancing transit agencies' responsiveness by extracting a broader range of information.
\end{abstract}

\maketitle
\section{Introduction}
Understanding user feedback is crucial for public transit agencies. Collecting and analyzing this feedback improves service quality, improves operational efficiency, builds community trust, and supports informed, data-driven decisions. Through user feedback, agencies gain insights into passengers' perceptions of the transit service, identifying satisfactory and unsatisfactory aspects, pinpointing when and where issues occur, and understanding overall needs. This process enables agencies to adapt to changing demands, address safety concerns, and maintain transparency and accountability. Ultimately, this leads to increased ridership and higher customer satisfaction.

However, collecting user feedback is a challenging process. Traditional methods, such as physical or online surveys, while effective, have several limitations. These methods have a limited reach and can be costly, requiring significant investments in money, time, and labor. In addition, respondents often experience survey fatigue, which can reduce the quality of feedback. Feedback collection through surveys is typically delayed, often conducted annually or subannually, which is useful for long-term analysis and decision making but impractical for understanding user perceptions in real-time~\cite{li2016better}.

Periodic surveys tend to capture opinions about the overall system or a broad network, making it difficult for agencies to obtain specific information about when, where, and what issues occurred. This lack of timely, detailed feedback hampers the ability of transit agencies to address immediate concerns and make prompt improvements to their services.

In contrast, collecting feedback through social media posts offers a more dynamic and cost-effective and relatively more real-time alternative~\cite{liu2017measuring}. Social media platforms allow agencies to reach a broader audience and collect crowd-sourced information from diverse users. This method is not only cost-effective, but also provides near-real-time feedback, which is crucial for timely responses and adjustments. Furthermore, social media offers access to a massive dataset, which can be seen as the foundation of applying complex analysis tools~\cite{schweitzer2012we}.

However, using social media data for user feedback analysis presents several challenges~\cite{grant2015enhancing}. From a data point of view, social networks often contain irrelevant or low-quality information. The linguistic structure is complex, with frequent use of sarcasm, slang, and informal language, making accurate interpretation difficult. Furthermore, social media data is less structured compared to traditional survey data, which lacks preset questions and predefined problem categories. Consequently, social media data related to public transit can be less specific, posing challenges to extract actionable information and increasing data labeling costs. Moreover, the high volume of data generated on social media platforms requires effective data management strategies to handle and analyze it efficiently.

From a methodological tool perspective, traditional Natural Language Processing (NLP) analysis tools for customer satisfaction often rely on fixed lexicons, limiting their ability to accurately interpret diverse language use on social media. As addressed in~\cite{kuflik2017automating}, social media posts, like tweets, are typically brief and lack context, making it difficult for traditional NLP methods to understand the full scope of an issue. Misinterpretations are common, especially when dealing with sarcasm, irony, or domain-specific information, without additional context.

In this paper, our objective is to improve the use of social media data for public transit user feedback by developing an advanced information extraction tool utilizing a Large Language Model (LLM). This tool offers improved sentiment analysis and expands the scope of feedback by moving beyond predefined topic categories, allowing for a more comprehensive understanding of user experiences. In addition, the tool is designed to extract geographical data from tweets, providing valuable location-specific insights. By increasing the ability to efficiently identify and address customer needs, this tool aims to enable more timely and effective service improvements, ultimately enhancing public transit services and boosting customer satisfaction.

The rest of this paper is organized as follows. The next section provides a general introduction to LLMs, discussing their capabilities and advancements compared with traditional NLP models. The following section explores state-of-the-art applications in the domain of social media analysis. Subsequently, the methodology and framework for using LLMs in the analysis of transit posts on social media are presented. The following section discusses the experimental results of using LLM for social media analysis. The last section addresses limitations, explores potential solutions, and presents possible future works.

\section{Literature Review}

This section provides a brief overview of widely used methods for information extraction from social media.

\subsection{Manual Identification}
Manual identification is a common method used to extract useful information from social media. For instance, authors in~\cite{osorio2021social} and ~\cite{lucini2020text} manually identified geo-information from tweets mentioning station or route names. Although this method guarantees accuracy, it is inefficient for large volumes of tweets. In~\cite{casas2017tweeting}, 1,624 tweets were identified for route information extraction, while~\cite{osorio2021social} involved manually labeling 3,454 tweets for spatial information. While human labeling is valuable for building benchmark datasets or lexicons, it is highly costly and impractical for frequent data analysis on fast-updating social media datasets.

\subsection{Lexicon-Based Approaches}
One of the most prominent methods for information extraction is the lexicon-based approach. This method identifies information, such as sentiment, using a collection of tokens with predefined scores. The overall sentiment of a sentence is determined based on the scores of these tokens~\cite{kiritchenko2014sentiment}. Lexicon-based approaches are common in sentiment analysis, where tokens are assigned scores of -1, 0, or 1, representing negative, neutral, or positive sentiments, respectively. The sentence score is then calculated accordingly~\cite{el2019linking, collins2013novel}. To enhance the performance of lexicon-based methods, it is often necessary to generate specific lexicons tailored to particular classification tasks. For instance, in~\cite{hosseini2018supporting}, a lexicon-based method was used to identify nine problematic topics in a tweet dataset related to the Calgary transit system. Similarly,~\cite{osorio2021social} employed 435 predefined terms to classify tweets into four topics: punctuality, comfort, breakdowns, and overcrowding. In another study,~\cite{al2024using} introduced a human-involved lexicon-building process to identify tweet topics, such as bus-related issues, sentiment, and sarcasm.

Despite its popularity, the lexicon-based approach has inherent drawbacks for social media data analysis. Firstly, dividing sentences into tokens can lead to the loss of contextual information. For example, irony or sarcasm can render individual tokens' meanings opposite to their intended sentiment in the sentence. Additionally, the informal writing habits prevalent on social media make it challenging to build comprehensive lexicon datasets that accommodate informal word usage.

\subsection{Machine Learning-Based Approaches}

Another notable method for information extraction is the machine learning (ML)-based approach. In this method, ML or deep learning (DL) models, such as logistic regression (LR), support vector machines (SVM), and decision trees (DT)~\cite{wankhade2022survey}, are trained on well-structured datasets for specific topic classification tasks. These models are commonly used in sentiment analysis. Additionally, transformer models with large structures, such as BERT, have been applied to topic classification tasks for social media analysis~\cite{osorio2021social}. A well-constructed dataset with relevant labels and similar content is crucial for ML-based methods. Transfer learning can alleviate this limitation by using models trained on datasets built for similar tasks in different domains, then applying them to current tasks~\cite{meng2019cross}. For example, in~\cite{leong2024metroberta}, the author used BERT to build a model for extracting transit topics from social media. Similarly, in~\cite{das2023classifying}, the BERT model was used for classiy pedestrian maneuver types. The model was trained on an online customer feedback dataset with 11 different topics and then used for further classification tasks related to the Washington Metropolitan Area Transit Authority.

However, even with transfer learning, ML-based approaches are still constrained by the scope of the training dataset and cannot identify topics not present in the dataset. Moreover, ML-based classification tasks are typically single-aspect. For instance, a model trained for sentiment analysis only performs sentiment analysis. This approach becomes costly when multiple aspects of information need to be extracted from the target text.

\subsection{Large Language Model-Based Approaches}
Large Language Models (LLMs) are a type of deep learning model built using multi-layer Transformer architectures~\cite{vaswani2017attention}, containing vast numbers of parameters and typically pre-trained on large-scale corpora~\cite{lu2029}. Through pre-training, LLMs acquire general knowledge, common sense, and the ability to understand and generate text. Due to their exceptional capabilities, LLMs are increasingly adopted for information extraction tasks. Beyond traditional natural language processing (NLP) tasks such as sentiment analysis~\cite{zhang2023sentiment} and semantic analysis~\cite{xu2024reasoning}, LLMs are widely applied in downstream tasks like knowledge reasoning, question answering, relation extraction, and event extraction, often outperforming traditional NLP models~\cite{xu2023large}.

In the transportation domain, LLMs have been used to answer transportation-related questions such as those concerning transportation economics and driver characteristics~\cite{syed2024benchmarking}. Additionally, LLMs are applied in areas such as transportation infrastructure planning and design~\cite{roberts2023gpt4geo}, project management~\cite{abbas2023relationship}, operations and maintenance~\cite{rane2023multidisciplinary}, and safety control~\cite{du2024large}. Multimodal LLMs are also used for tasks like object detection in transportation~\cite{ashqar2024advancing}.

As noted in~\cite{papageorgiou2024survey, tupayachi2024towards}, high-quality data ecosystems are essential for effective LLM applications in specific domains. For instance, efforts to benchmark LLM performance in transportation have resulted in datasets like TransportBench~\cite{syed2024benchmarking}, designed to assess reasoning abilities in transportation problems, and various question-answer datasets~\cite{zhang2023study}, which evaluate decision-making, complex event causality reasoning, and human driving exam performance. Moreover,~\cite{prajapati2024evaluation} provides an image dataset for evaluating LLMs' ability to detect transportation-related issues like cracks or congestion. The GTFS-related dataset in~\cite{devunuri2024chatgpt} evaluates LLM performance in semantic understanding and information retrieval.

However, as discussed in~\cite{du2024large}, while LLMs are powerful, they face challenges when applied to domain-specific tasks due to knowledge gaps. Fine-tuning is a potential solution, as demonstrated in~\cite{wang2024transgpt}, where an open-source multimodal LLM, VisualGLM, was fine-tuned with 12.5 million textual tokens to enhance its transportation domain knowledge. Similarly, in~\cite{zheng2023trafficsafetygpt}, LLaMA 3 was fine-tuned with a traffic safety dataset to improve its performance in that domain.

Fine-tuning pre-trained LLMs requires large amounts of high-quality training data and significant computational resources. A lightweight alternative is the Retrieval-Augmented Generation (RAG) system, which retrieves domain-specific information from external databases, as proposed in~\cite{xu2024genai}. However, this framework remains largely conceptual, with limited real-world applications.

To address the gap in applying LLMs to transportation information extraction, particularly in user feedback analysis, we introduce Transit Pulse. This is one of the first attempts to provide a lightweight solution for semi-automatic information extraction from social media posts, offering insights for transit monitoring and control.

\section{Methodology}

This paper addresses two user feedback analysis tasks: traditional classification and information extraction \& summarizing. For the traditional multi-class classification task, we cover sentiment classification, sarcasm detection, and transit problem topic classification. This section introduces the classification and information extraction pipelines used in our experiments.

\subsection{Traditional Classification Task}

\begin{figure}[pos=!htbp]
    \centering
    \includegraphics[width=1\linewidth]{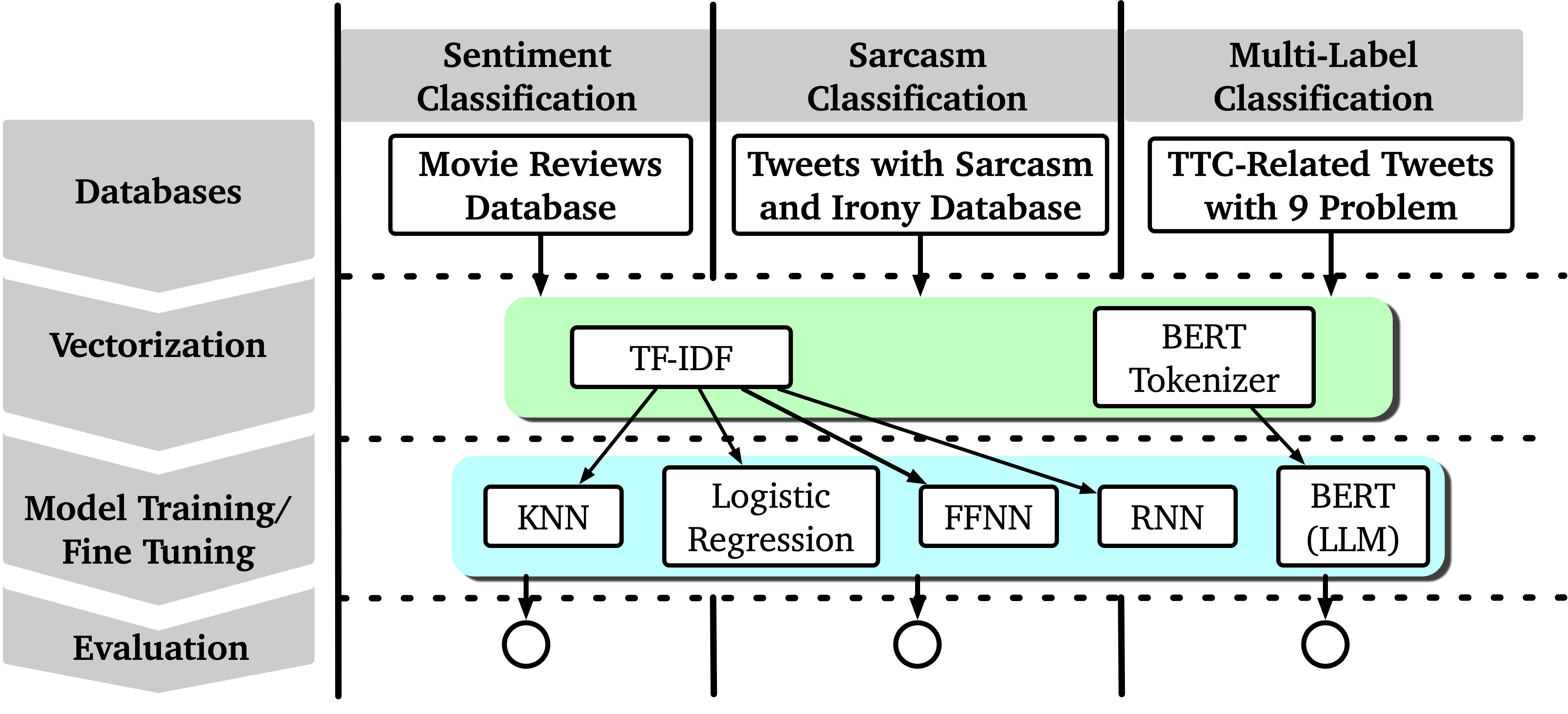}
    \caption{Framework for the Traditional Classification Task, illustrating the steps from data processing to model evaluation.}
    \label{fig:ctf}
\end{figure}

As shown in Fig.~\ref{fig:ctf}, the traditional classification process involves data processing, vectorization, model training or fine-tuning, and model evaluation. We use both traditional NLP methods---Term Frequency-Inverse Document Frequency (TF-IDF) with machine learning (ML)-based classification---and large language model (LLM)-based classification methods.

The initial step involves vectorizing or tokenizing the neutral language input using either TF-IDF or the BERT tokenizer. This converts the input into a multi-dimensional space vector for further training.

\subsubsection{TF-IDF}

TF-IDF is a statistical measure that evaluates the importance of a word in a document relative to a collection of documents (corpus). It is widely used in information retrieval and text mining.

\begin{itemize}
    \item \textbf{Term Frequency (TF)} measures how frequently a term occurs in a document, normalized to prevent bias towards longer documents:
    \[
    \textit{tf}(t, d) = \frac{\text{Number of times term } t \text{ appears in document } d}{\text{Total number of terms in document } d}
    \]
    \item \textbf{Inverse Document Frequency (IDF)} measures the importance of a term across the corpus, reducing the weight of frequently occurring terms:
    \[
    \textit{idf}(t, D) = \log \left( \frac{N}{|\{d \in D : t \in d\}|} \right)
    \]
    where \(N\) is the total number of documents in the corpus \(D\), and \(|\{d \in D : t \in d\}|\) is the number of documents in which the term \(t\) appears.
    \item \textbf{TF-IDF score} for a term \(t\) in a document \(d\) is the product of its term frequency and inverse document frequency:
    \[
    \textit{tf-idf}(t, d, D) = \textit{tf}(t, d) \times \textit{idf}(t, D)
    \]
\end{itemize}

TF-IDF provides a numerical statistic reflecting the importance of a word to a document in a corpus, enabling effective text analysis and retrieval.

\subsubsection{BERT Tokenizer and BERT Model}

In contrast to TF-IDF, the BERT tokenizer, part of a pre-trained LLM (the embedding layer in BERT), tokenizes and embeds text into dense vectors with contextual meaning. The BERT tokenizer splits text into tokens using WordPiece, then processes these tokens through bidirectional transformers pre-trained on large corpora to capture contextual information.

The tokenized input is further processed by the pre-trained BERT model~\cite{devlin2018bert}, which uses self-attention to learn contextual information. The BERT model, thanks to its bidirectional structure, understands word meanings from both directions in a sentence and from distant words. Fine-tuning the BERT model for classification involves adding a dense layer that takes the model’s output as a high-level feature vector for final classification. Pre-trained on massive datasets, the BERT model’s general understanding of language aids the classification layer in making accurate classifications.

\subsection{Information Extraction}

Traditional classification methods are easy to deploy and generally reliable but face limitations. They are restricted by the quality and scope of the training data set and can only classify predefined topics. Models trained for specific tasks, such as sentiment classification, cannot be directly used for other tasks like sarcasm detection. Additionally, performance can degrade when applied to scenarios different from the training dataset.

To overcome these limitations, we introduce an information extraction pipeline based on the powerful open-source LLM, Llama 3 by Meta~\cite{llama3modelcard}. Llama 3 excels in over 150 benchmark tasks, including answering science / domain-specific questions and common sense reasoning~\cite{open-llm-leaderboard}. Its power comes from high-quality training data and a large model structure. Llama 3 was pre-trained with more than 15 trillion tokens, covering all high-quality open data available until December 2023. The model’s 70 billion parameters enable it to learn from this massive dataset through supervised fine-tuning (SFT) and reinforcement learning with human feedback (RLHF), resulting in accurate and human-preferred outputs.

\begin{figure}[pos=!htbp]
    \centering
    \includegraphics[width=1\linewidth]{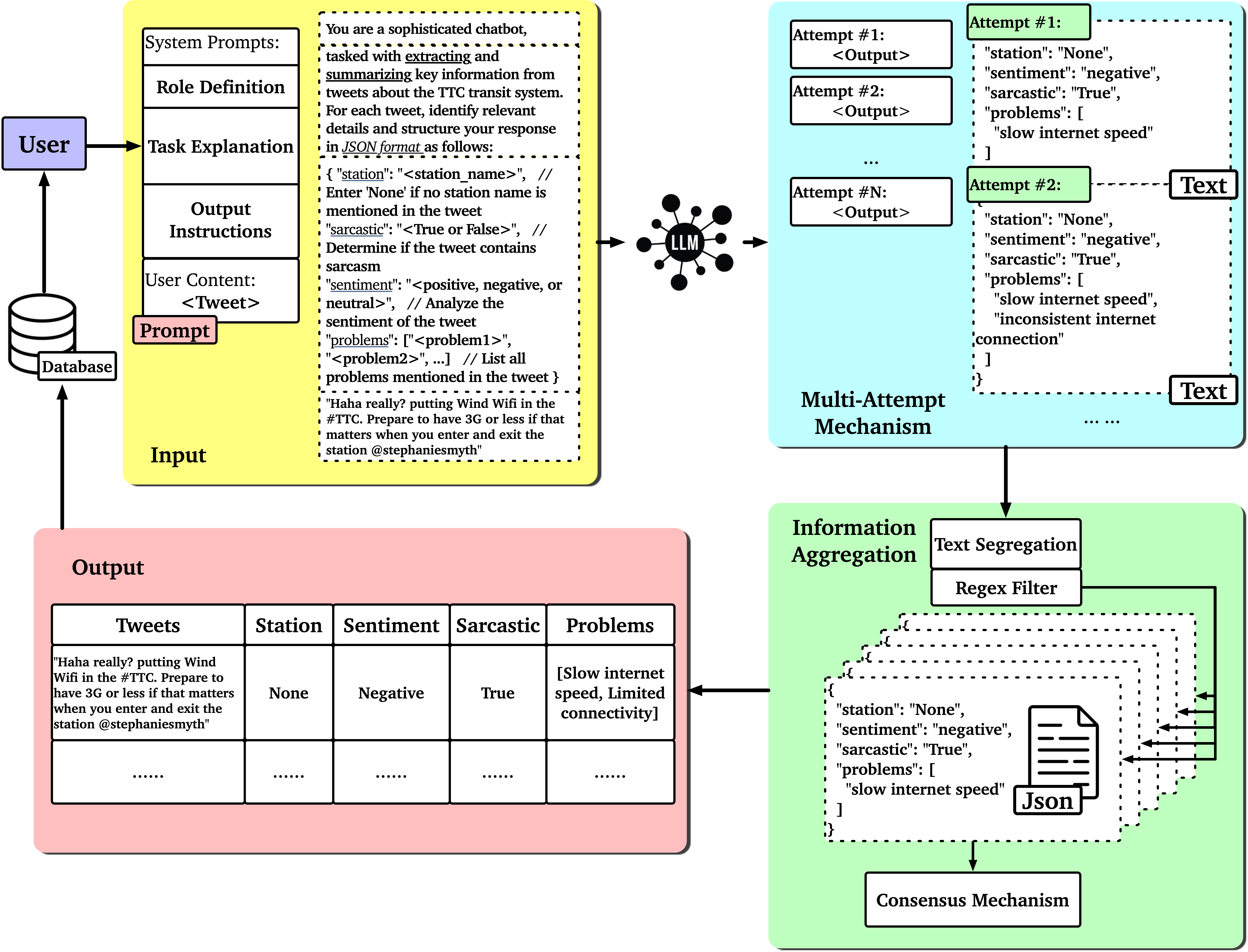}
    \caption{Information Extraction Pipeline with Llama 3, demonstrating the process of embedding tweets into structured prompts and aggregating the extracted information.}
    \label{fig:infoextractpip}
\end{figure}

As illustrated in Fig.~\ref{fig:infoextractpip}, the information extraction pipeline begins by embedding target tweets into a structured prompt to guide the LLM in extracting the desired information. The prompt defines the LLM’s role and tasks, including extracting and summarizing information related to the transit agency tweeted about (in our case the Toronto Transit Commission, TTC for short), such as station name, sentiment, sarcasm, and problem topic. The output is JSON-like text, which is processed through an information aggregation step. This step involves segregating text chunks into key-value pairs, filtering unuseful text with regex, and constructing a structured JSON dataset. The consensus mechanism determines the most common answer from multiple LLM responses to mitigate performance variations.

\subsection{Retrieval Augmented Generation (RAG) System}

\begin{figure}[pos=!htbp]
    \centering
    \begin{subfigure}[pos=t]{0.9\linewidth}
        \centering
        \includegraphics[width=\linewidth]{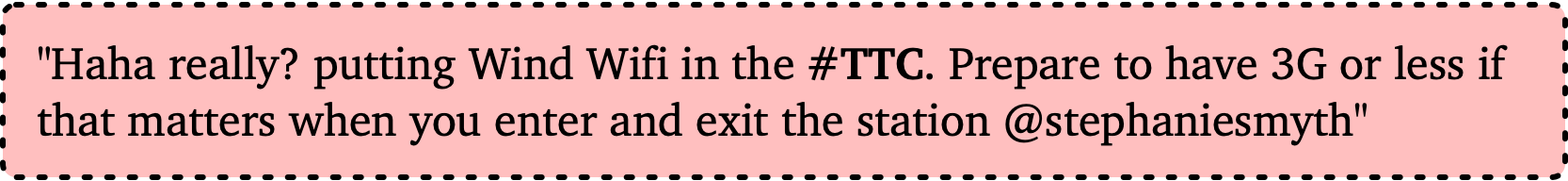}
        \caption{Extracted "TTC" as station name.}
        \label{fig:station_wrong_exp1}
    \end{subfigure}
    \begin{subfigure}[t]{0.9\linewidth}
        \centering
        \includegraphics[width=\linewidth]{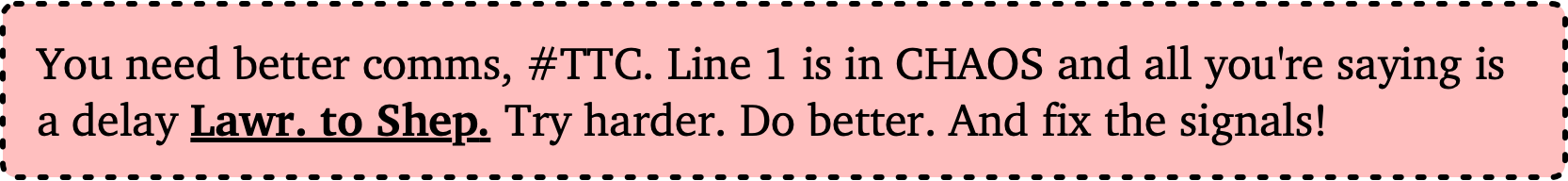}
        \caption{Extracted "Lawr and Shep" as station name.}
        \label{fig:station_wrong_exp_2}
    \end{subfigure}
    \begin{subfigure}[t]{0.9\linewidth}
        \centering
        \includegraphics[width=\linewidth]{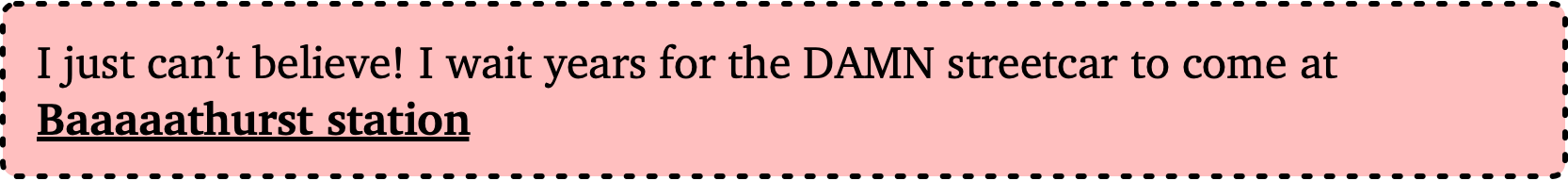}
        \caption{Extracted "Baaaaaathurst Station" as station name.}
        \label{fig:station_wrong_exp_3}
    \end{subfigure}
    \caption{Examples of Station Information Misinterpretation in Tweets.}
    \label{fig:station_wrong_exp}
\end{figure}

Despite Llama 3's strengths in context understanding and NLP tasks, it has limitations with unfamiliar data or domain-specific knowledge. For instance, it may misinterpret station names in TTC-related tweets, especially when names are abbreviated or misspelled, as shown in Fig.~\ref{fig:station_wrong_exp}.

To address this, we implemented a Retrieval Augmented Generation (RAG) system~\cite{lewis2020retrieval}. The RAG system supplements the LLM with external knowledge to improve accuracy in domain-specific information extraction.

\begin{figure}[pos=!htbp]
    \centering
    \includegraphics[width=0.9\linewidth]{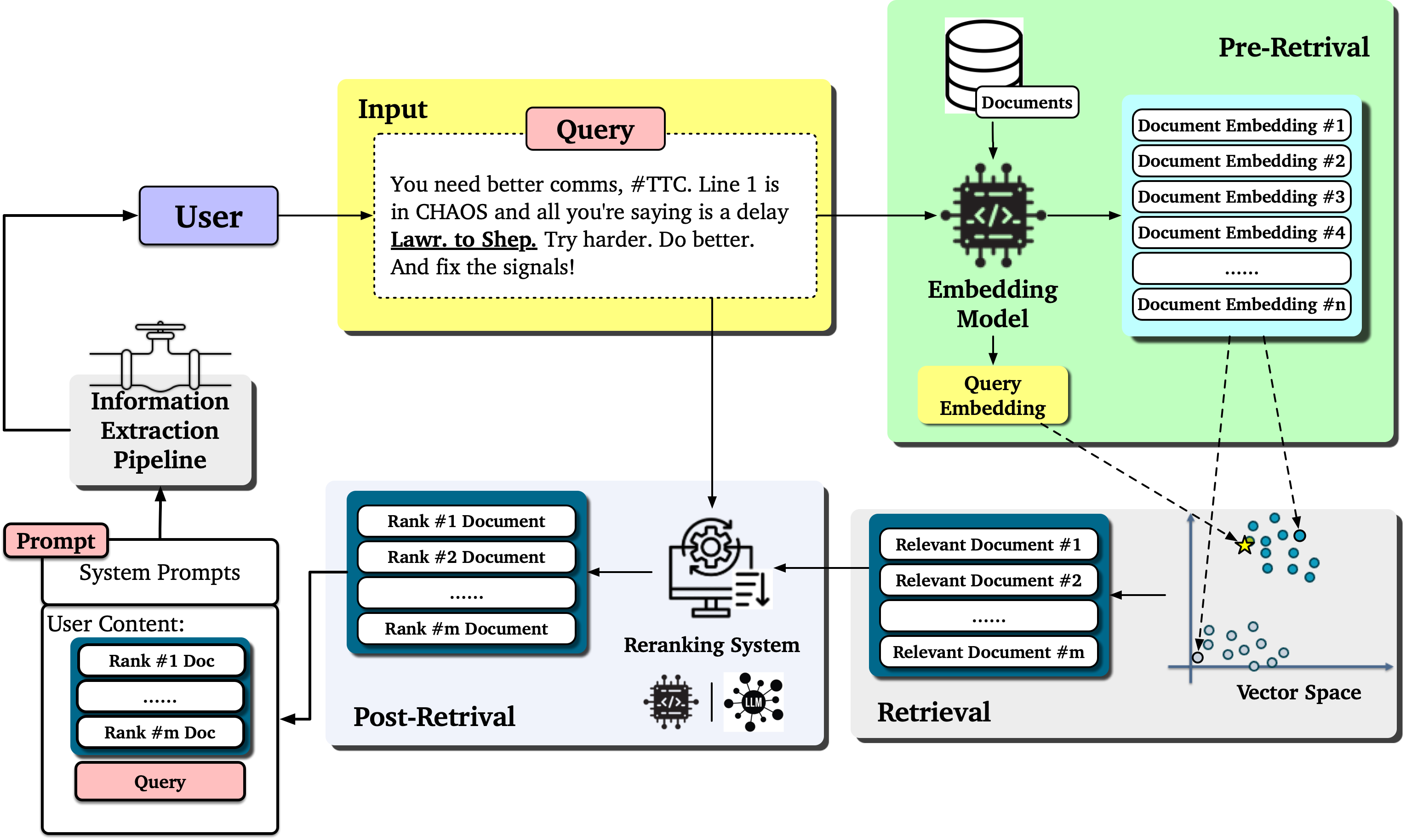}
    \caption{Retrieval Augmented Generation (RAG) System, showing the process of embedding external knowledge and matching it with user queries to enhance information extraction accuracy.}
    \label{fig:rag}
\end{figure}

As shown in Fig.~\ref{fig:rag}, the RAG process starts by embedding the external knowledge base using a pre-trained LLM embedding model in a vector space. Each document is represented as a point in this high-dimensional space. When a query (e.g., tweet content) is embedded into the same space, we compare it to the knowledge base to find the closest matches. These matches are relevant external documents selected based on distance metrics like Euclidean Distance, cosine similarity, or maximum inner product (MIP). The retrieval process typically outputs more candidates than needed for re-ranking by a more complex embedding system or LLM-driven prompts. The retrieved information is then added to the information extraction pipeline, enhancing accuracy and comprehensiveness.

In summary, our methodology combines advanced LLM techniques and the RAG system, to effectively classify and extract information from social media data related to public transit.

\section{Experiments}

This section presents our experiments on using LLM-based pipeline for traditional classification problems and information extraction \& summarization. We begin by introducing the experimental setup, including datasets, models, and hardware configuration. Then, we discuss model performance on each dataset and the information extracted by our pipeline. Additionally, we provide a case study showcasing the practical application of our information extraction system for real-time public transit service monitoring.

\subsection{Datasets}

We used three datasets to test model performance on multi-class classification tasks for sentiment analysis, sarcasm detection, and transit problem classification.

For sentiment analysis, we used the Sentiment Analysis on Movie Reviews dataset~\cite{sentiment-analysis-on-movie-reviews}, containing 156,060 training records and 66,292 testing records across five sentiment groups: negative, somewhat negative, neutral, somewhat positive, and positive.

For sarcasm detection, we used the Tweets with Sarcasm and Irony dataset~\cite{ling2016empirical}, which includes four classes: irony, sarcasm, regular, and figurative. The dataset has 54,618 training records and 7,861 testing records.

These datasets are publicly available and specifically designed for sentiment analysis and sarcasm detection. We did not use our TTC tweets dataset for performance testing due to the lack of human-labeled sentiment and sarcasm information.
\begin{figure}[pos=htpb!]
    \centering
    \includegraphics[width=0.8\linewidth]{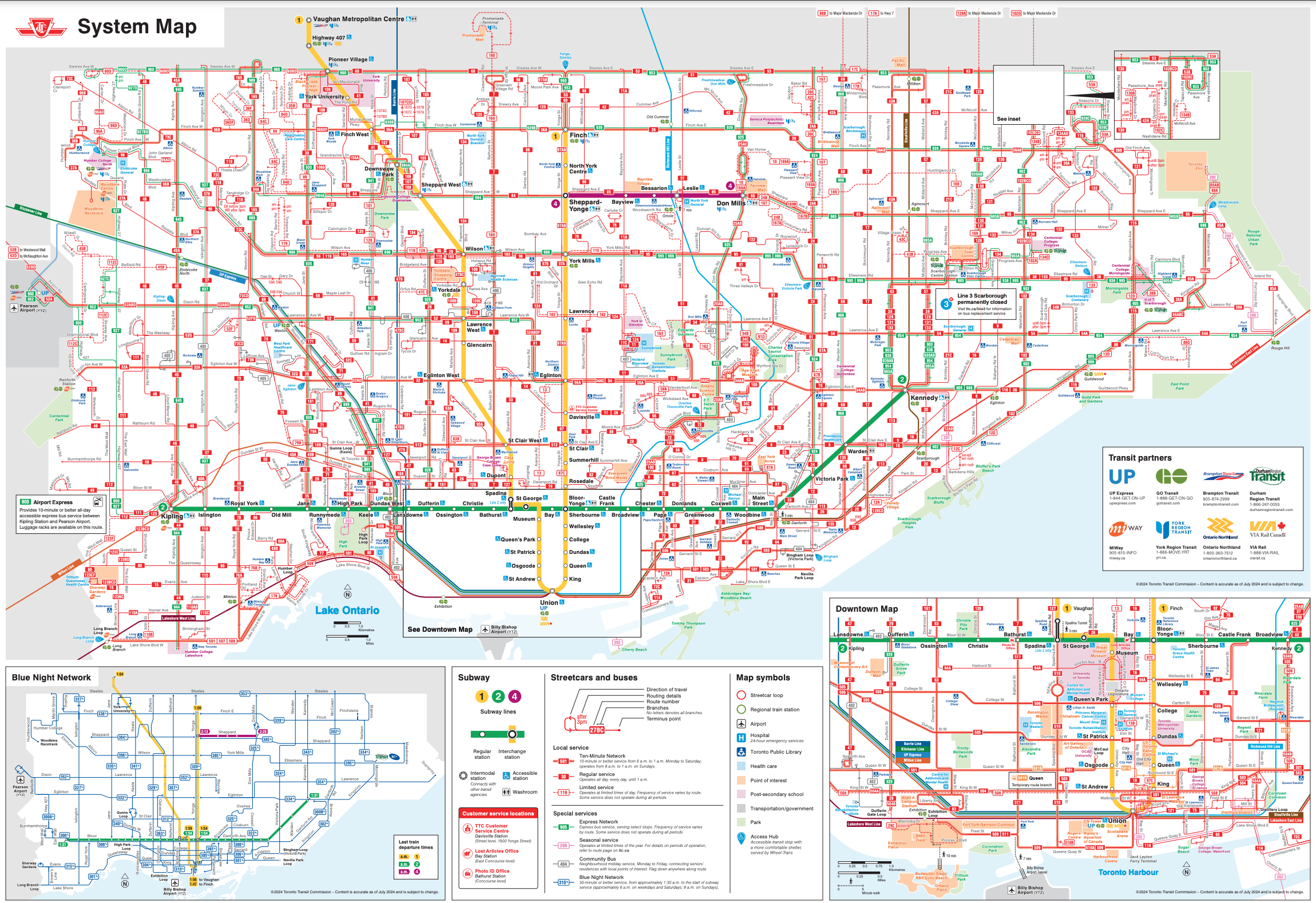}
    \caption{System map of the Toronto Transit Commission (TTC)~\cite{TTC_map}.}
    \label{fig:ttc_map}
\end{figure}

\begin{figure}[pos=!htbp]
    \centering
    \begin{subfigure}[t]{0.49\linewidth}
        \centering
        \includegraphics[width=\linewidth]{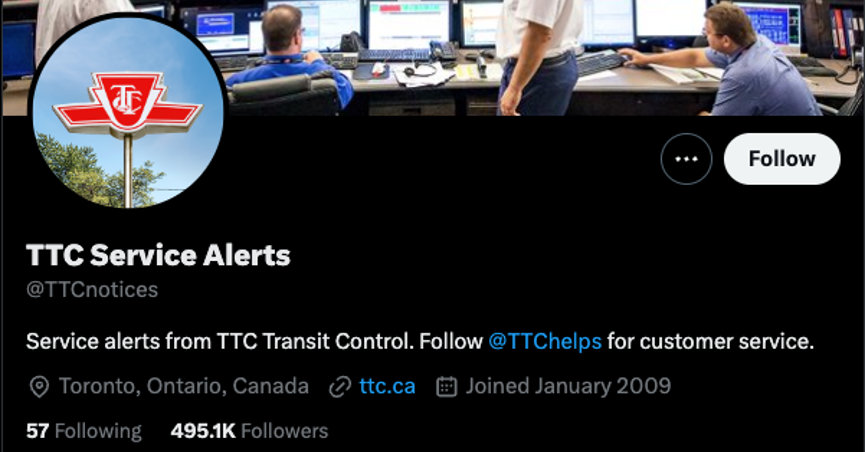}
        \caption{TTC Service Alerts, 495.1 thousand followers when taking this screenshot}
    \end{subfigure}
    \hfill
    \begin{subfigure}[t]{0.49\linewidth}
        \centering
        \includegraphics[width=\linewidth]{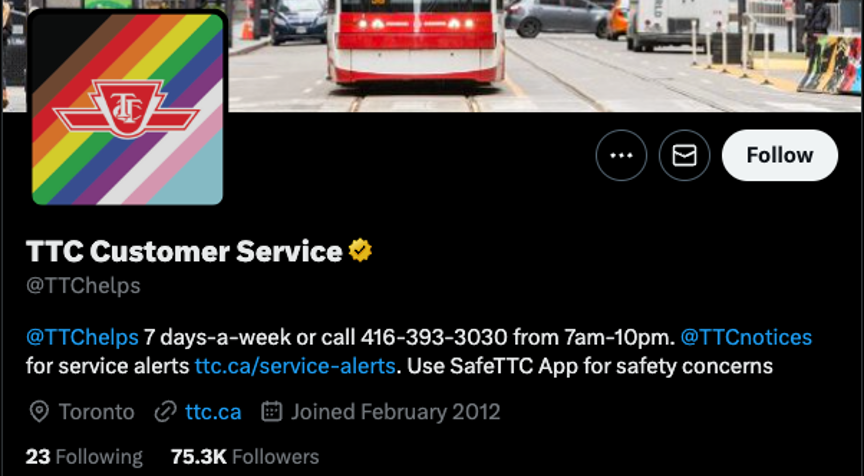}
                \caption{TTC Customer Service, 75.3 thousand followers when taking this screenshot}
    \end{subfigure}
    \caption{Two Main Official X (formerly Twitter) Accounts of TTC (Screenshot taken on June 26, 2024).}
    \label{fig:t_ttc_accounts}
\end{figure}

For the transit-related problem classification task, we used tweets related to the Toronto Transit Commission (TTC), which also serves as our information extraction dataset. The data was collected from February 5th, 2015, to December 31st, 2015, from two main TTC X (formerly Twitter) accounts (as shown in Fig.~\ref{fig:t_ttc_accounts}): TTC Service Alerts and TTC Customer Service. The TTC operates 192 bus routes, 11 streetcar routes, and 3 subway lines, serving over 1.4 million riders daily at its peak in 2023~\cite{TTC_fact}.

\begin{figure}[pos=!htbp]
    \centering
    \includegraphics[width=0.5\linewidth]{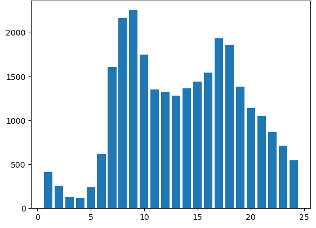}
    \caption{Number of tweets published at different times of the day.}
    \label{fig:tdtd}
\end{figure}

The TTC-related tweets dataset includes 631,691 records. After removing duplicates and retweets with minimal additional information, 27,312 tweets remained for analysis. Figure~\ref{fig:tdtd} shows the number of tweets posted at different times of the day, peaking during peak hours.

The dataset categorizes tweets into 10 problem categories: maintenance, capacity availability, interaction with staff, travel time, ride quality, winter maintenance, temporal availability, safety and security, accessibility, and communication (Table~\ref{tab:transit_aspects}). Initially, a group of keywords for each problem topic was manually identified from a subset of the entire dataset. Subsequently, a lexicon-based method was employed to label the remaining tweets based on the frequency of keyword occurrences, assigning each tweet to the most relevant problem category.
\begin{table}
\centering
\begin{tabular}{l>{\centering\arraybackslash}p{3cm}>{\centering\arraybackslash}p{3cm}}
\toprule
\textbf{Category} & \textbf{Count} \\
\midrule
Winter Maintenance & 24 \\
Temporal Availability & 179 \\
Interaction with Staff & 245 \\
Maintenance & 396 \\
Capacity Availability & 553 \\
Communication & 648 \\
Accessibility & 1263 \\
Ride Quality & 1599 \\
Travel Time & 2882 \\
Safety and Security & 2943 \\
\bottomrule
\end{tabular}
\caption{Summary of various transit system aspects and their counts.}
\label{tab:transit_aspects}
\end{table}

\begin{figure}[pos=!htbp]
    \centering
    \includegraphics[width=0.8\linewidth]{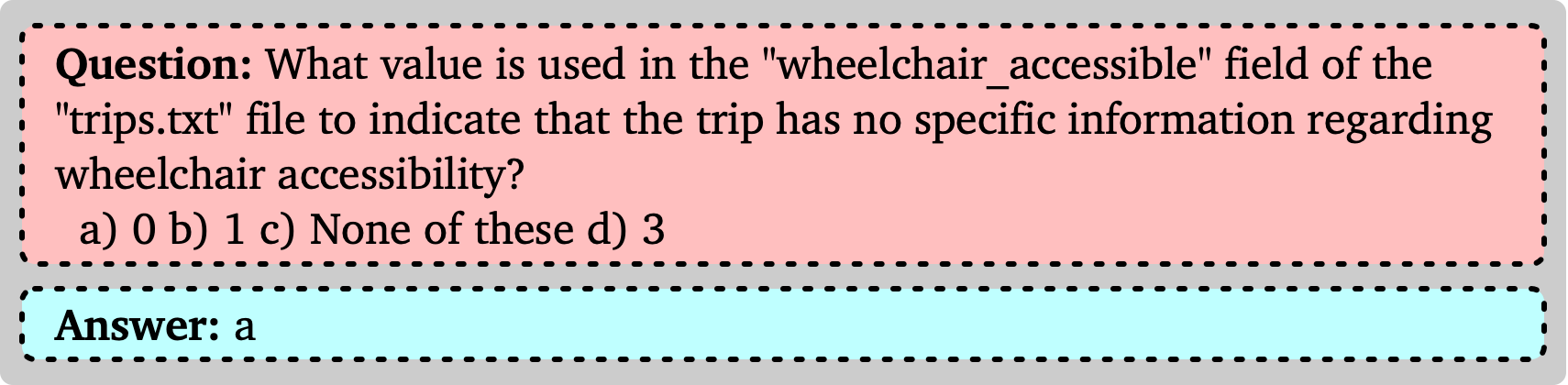}
    \caption{Example of GTFS understanding benchmarking questionnaire.}
    \label{fig:gtfs_q_e_u}
\end{figure}

To evaluate model performance with and without the RAG system, we utilized a GTFS-specific question dataset~\cite{devunuri2023chatgpt} designed to assess the understanding of GTFS standards and the retrieval of information from structured GTFS data. This dataset comprises 195 questions across six categories: term definitions, common reasoning, file structure, attribute mapping, data structure, and categorical mapping. An example of these questions is shown in Fig.~\ref{fig:gtfs_q_e_u}.

According to~\cite{devunuri2023chatgpt}, Term definitions questions test the model's ability to understand specific GTFS terms and document structures. Common reasoning questions evaluate basic GTFS knowledge, including abbreviations, usage, and file purposes. File structure questions determine the model's ability to identify the correct files in given contexts. Attribute mapping questions assess whether the model can correctly associate attributes with their respective files. Data structure questions verify the model's capability to identify attribute data types accurately. Categorical mapping questions examine the model's understanding of categorical variables representing data, such as different codes for wheelchair availability on buses.

\begin{figure}[pos=!htbp]
    \centering
    \includegraphics[width=0.8\linewidth]{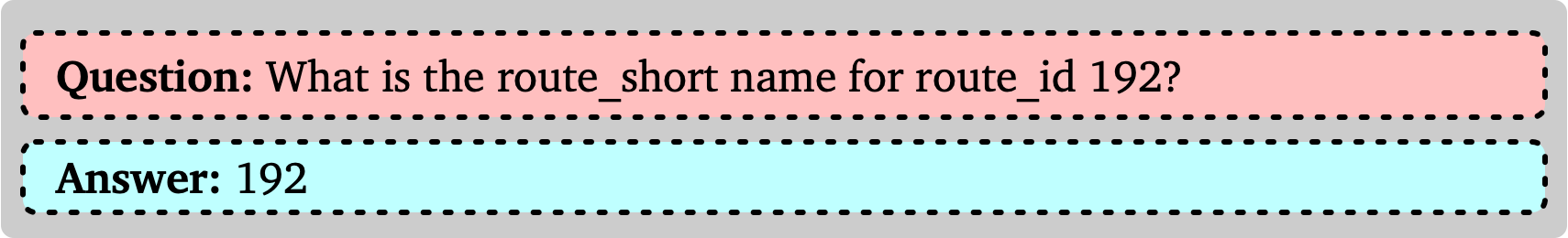}
    \caption{Example of GTFS programming question.}
    \label{fig:gtfs_t_prog}
\end{figure}

In addition to these, the dataset includes 87 programming questions aimed at testing the model's ability to retrieve information from the GTFS dataset effectively. An example of these questions is shown in Fig.~\ref{fig:gtfs_t_prog}.

\subsection{Models and Environment Setup}

We implemented various models for classification tasks, including logistic regression, KNN, FFNN, and RNN, to compare with the relatively small LLM: BERT. As discussed in the methodology section, traditional classification methods using ML/DL are based on TF-IDF for text embedding and trained from scratch. In contrast, BERT is fine-tuned for each task using corresponding datasets, showcasing the power of LLM structures.

For the information extraction pipeline, we used LLama-3 as the core for text understanding, information extraction, and summarization.

The experiments were conducted in the following environment:

\begin{itemize}
    \item \textbf{System}: Pop OS
    \item \textbf{Processor}: 2.4 GHz 8-Core Intel Core i9
    \item \textbf{GPU}: Nvidia 4090
    \item \textbf{Programming Language}: Python
    \item \textbf{Large Language Model (for classification)}: BERT
    \item \textbf{Large Language Model (for information extraction)}: LLama-3 8B
    \item \textbf{Prompt Framework}: LangChain
\end{itemize}

\subsection{Experiment Results}

\subsubsection{Classification Tasks}

Table~\ref{table:performance} presents the accuracy results of various models across three different tasks: Sentiment Analysis, Sarcasm Detection, and Problem Topic Classification.

\begin{table}
\centering
\captionsetup{justification=centering}
\begin{tabular}{l>{\centering\arraybackslash}p{3cm}>{\centering\arraybackslash}p{3cm}>{\centering\arraybackslash}p{3cm}}
\toprule
\textbf{Model} & \textbf{Sentiment Analysis} & \textbf{Sarcasm Detection} & \textbf{Problem Topic Classification} \\
\midrule
Logistic Regression (TF-IDF) & 62.90\% & 74.32\% & 81.59\% \\
KNN & 63.19\% & 60.80\% & 84.40\% \\
FNN & 64.87\% & 75.46\% & 86.83\% \\
RNN & 64.71\% & 75.39\% & 79.12\% \\
BERT (LLM) & \textbf{78.02\%} & \textbf{82.64\%} & \textbf{91.50\%} \\
\bottomrule
\end{tabular}
\caption{Model Performance Comparison}
\label{table:performance}
\end{table}

The LLM model demonstrates the highest accuracy across all three tasks, substantially outperforming other models, especially in Sentiment Analysis and Sarcasm Detection. For Problem Topic Classification, BERT (LLM) achieves 91.5\% accuracy, highlighting its robustness in handling complex language understanding tasks compared to traditional and other neural network models.

\subsubsection{GTFS Understanding and Retrieval}

\begin{figure}[pos=!htbp]
    \centering
    \begin{subfigure}[t]{0.49\linewidth}
        \centering
        \includegraphics[width=\linewidth]{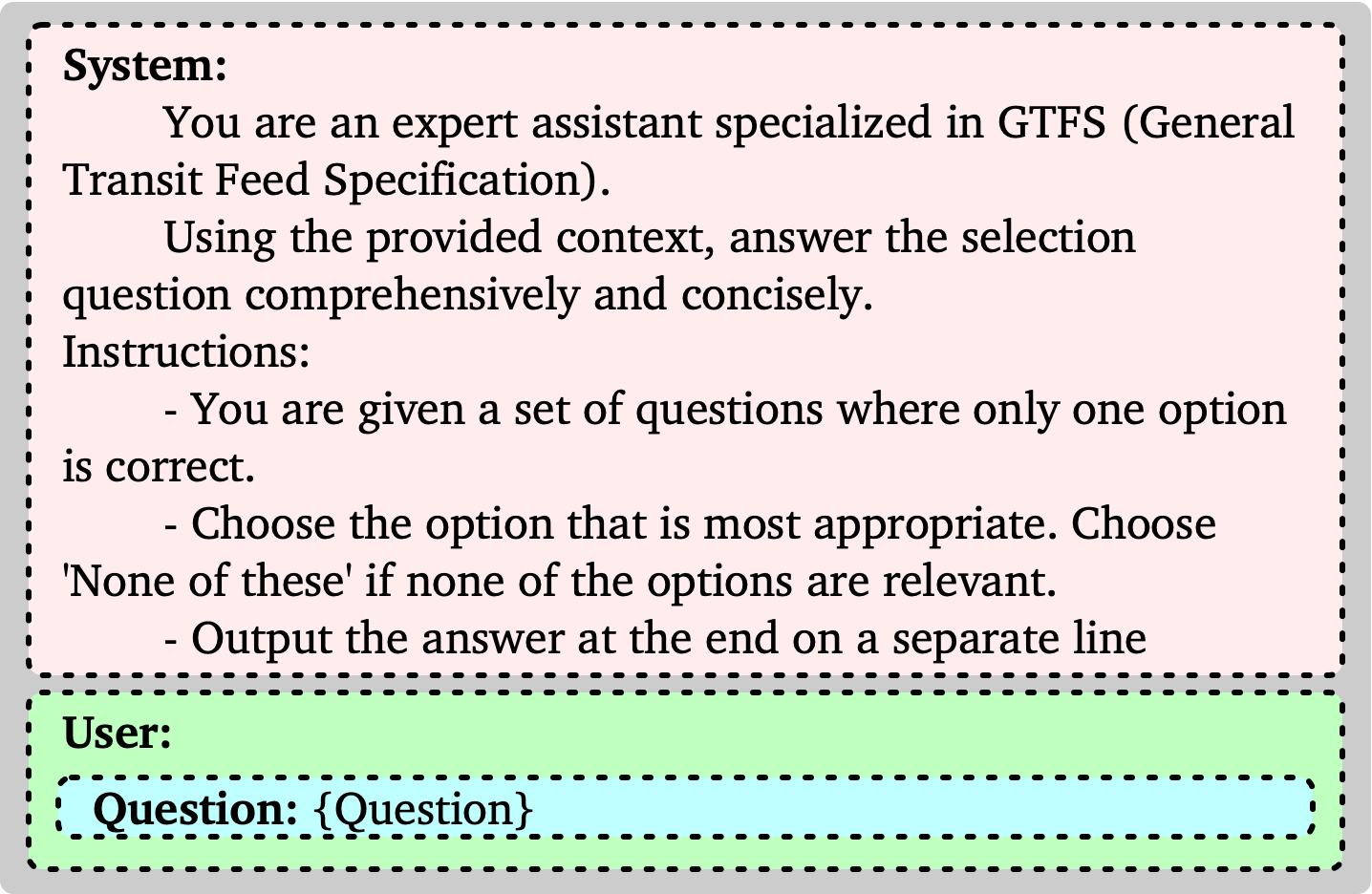}
        \caption{Prompt without retrieved information from RAG system}
    \end{subfigure}
    \hfill
    \begin{subfigure}[t]{0.49\linewidth}
        \centering
        \includegraphics[width=\linewidth]{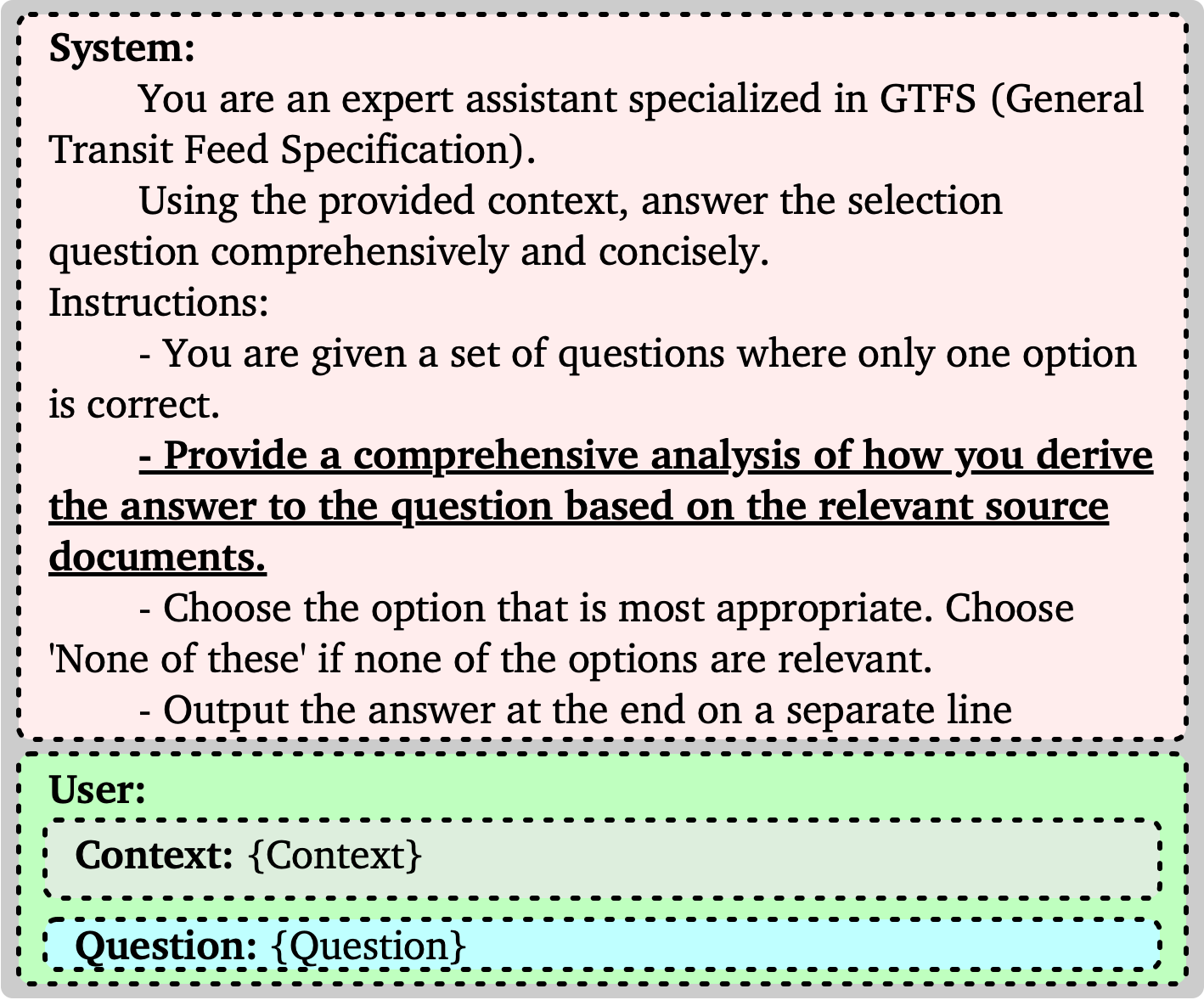}
        \caption{Prompt with retrieved information from RAG system}
    \end{subfigure}
    \caption{Prompts for LLM answering GTFS understanding multiple choice questions, with and without information from RAG system}
    \label{fig:rag_prompt}
\end{figure}

This section presents the performance of the LLM for GTFS understanding and retrieval tasks, with and without the RAG system. Figure~\ref{fig:rag_prompt} shows the prompts for the LLM answering multiple choice questions on GTFS understanding, both with and without the retrieved information from the RAG system. The external documents in the RAG system were built using the official GTFS documentation\footnote{\url{https://gtfs.org/schedule/reference/}}.

\begin{figure}[pos=!htbp]
    \centering
    \begin{subfigure}[t]{0.49\linewidth}
        \centering
        \includegraphics[width=\linewidth]{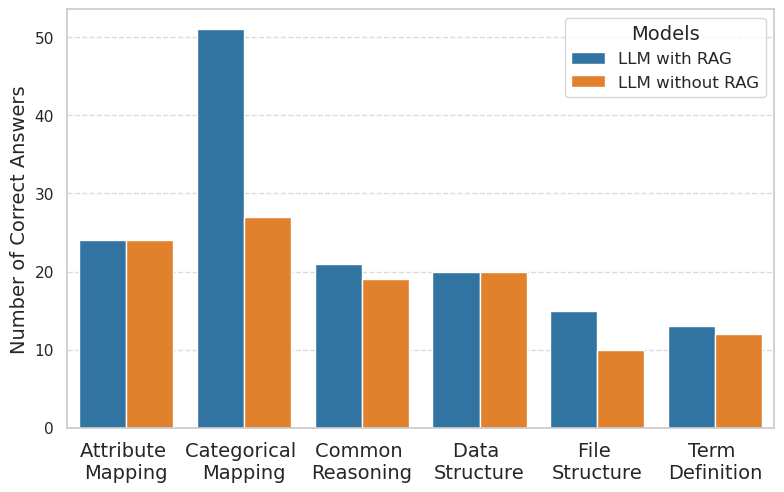}
        \caption{Correct answers for GTFS understanding questions with and without RAG}
    \end{subfigure}
    \hfill
    \begin{subfigure}[t]{0.49\linewidth}
        \centering
        \includegraphics[width=\linewidth]{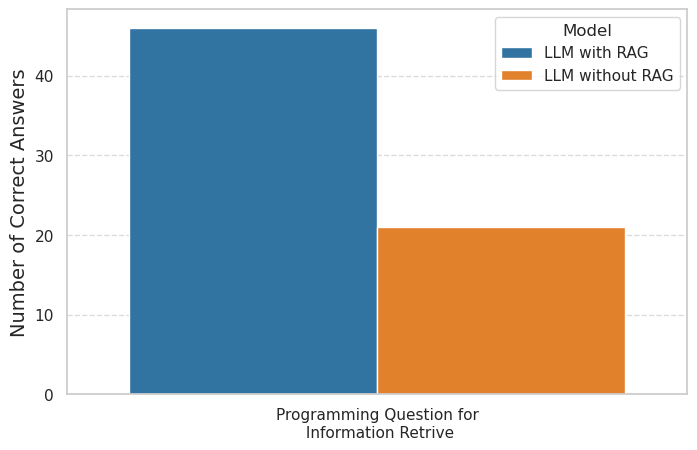}
        \caption{Correct answers for GTFS programming questions with and without RAG}
    \end{subfigure}
    \caption{Performance of LLM on GTFS tasks with and without RAG system}
    \label{fig:rag_gtfs_results}
\end{figure}

The results of these experiments are shown in Figure~\ref{fig:rag_gtfs_results}. For multiple choice questions across six types, the LLM with RAG consistently achieves a higher number of correct answers. Notably, for Categorical Mapping questions, the model with RAG scores 51 correct answers compared to 27 without RAG. Overall, the accuracy of answering GTFS-related multiple choice questions increases from 57.43\% to 73.85\%. For programming questions, accuracy improves from 24.14\% (21/87) to 52.87\% (46/87), demonstrating the enhanced capability of RAG in improving LLM performance for domain-specific questions.

\subsubsection{Station Name Extraction with RAG}

\begin{figure}[pos=!htbp]
    \centering
    \includegraphics[width=0.7\linewidth]{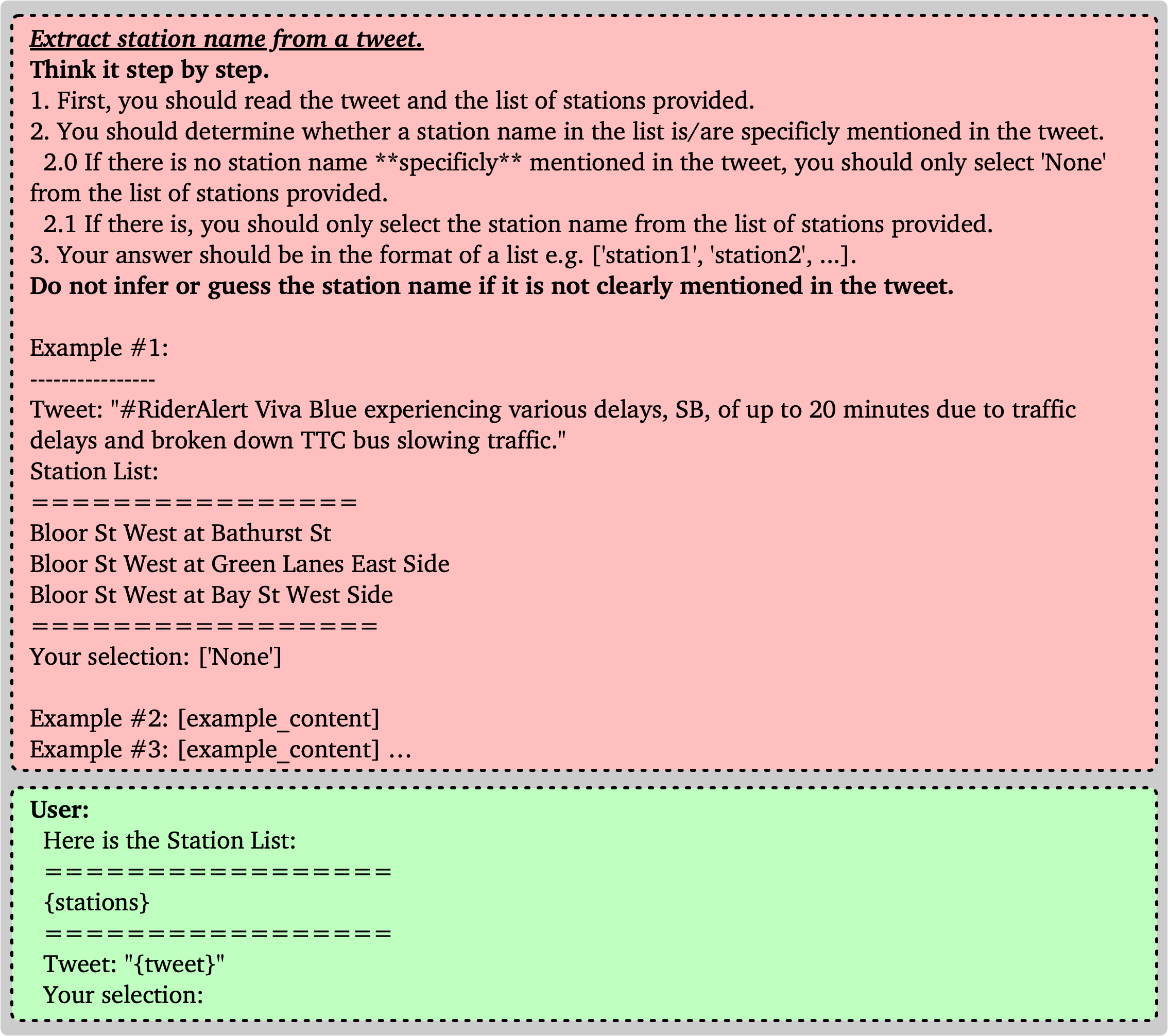}
    \caption{Prompts used in the RAG system to improve station name extraction.}
    \label{fig:rag_ttc_station}
\end{figure}

The prompts used in the RAG system during the post-retrieval phase to enhance the accuracy and formality of station name extraction are shown in Fig.~\ref{fig:rag_ttc_station}. To guide the language model in selecting the most relevant station name from the retrieved list, we employed the Chain-of-Thoughts (CoT) and Few-Shot techniques. These methods instruct the model to think step-by-step and provide reference examples for making decisions.

\begin{figure}[pos=!htbp]
    \centering
    \includegraphics[width=1\textwidth]{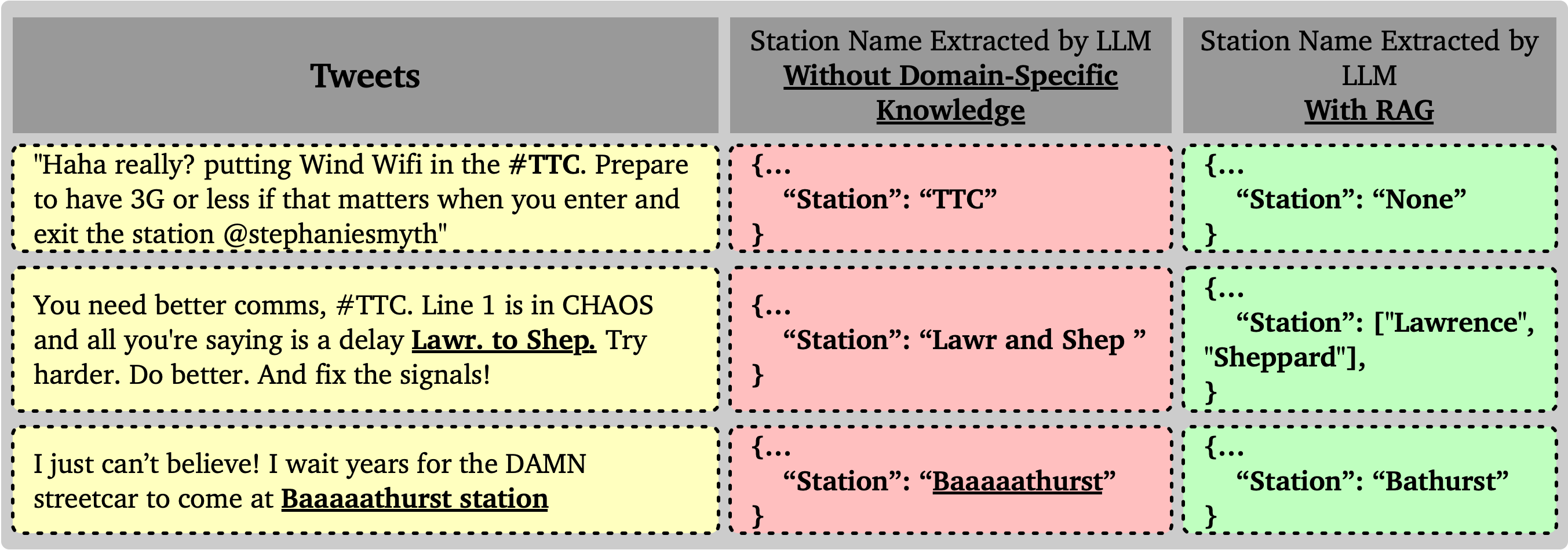}
    \caption{Comparison of station name extraction by Llama 3 with and without the RAG system.}
    \label{fig:station_rag_exp}
\end{figure}

Figure~\ref{fig:station_rag_exp} shows three examples of using the RAG system for station extraction. As illustrated, the RAG system enables the language model to extract more formal station information, avoiding issues related to using abbreviations or misspellings. Additionally, the model can better handle instances where "TTC" is incorrectly identified as a station name.

\subsubsection{Information Extraction with LLM}

This section demonstrates information extraction from the TTC-related tweets dataset using LLM, compared with traditional NLP methods. A notable advantage of the LLM is its ability to generate comprehensive outputs, including station information, sentiment, sarcasm, and problem summarization in a single response.

\begin{figure}[pos=htbp!]
    \centering
    \includegraphics[width=1\linewidth]{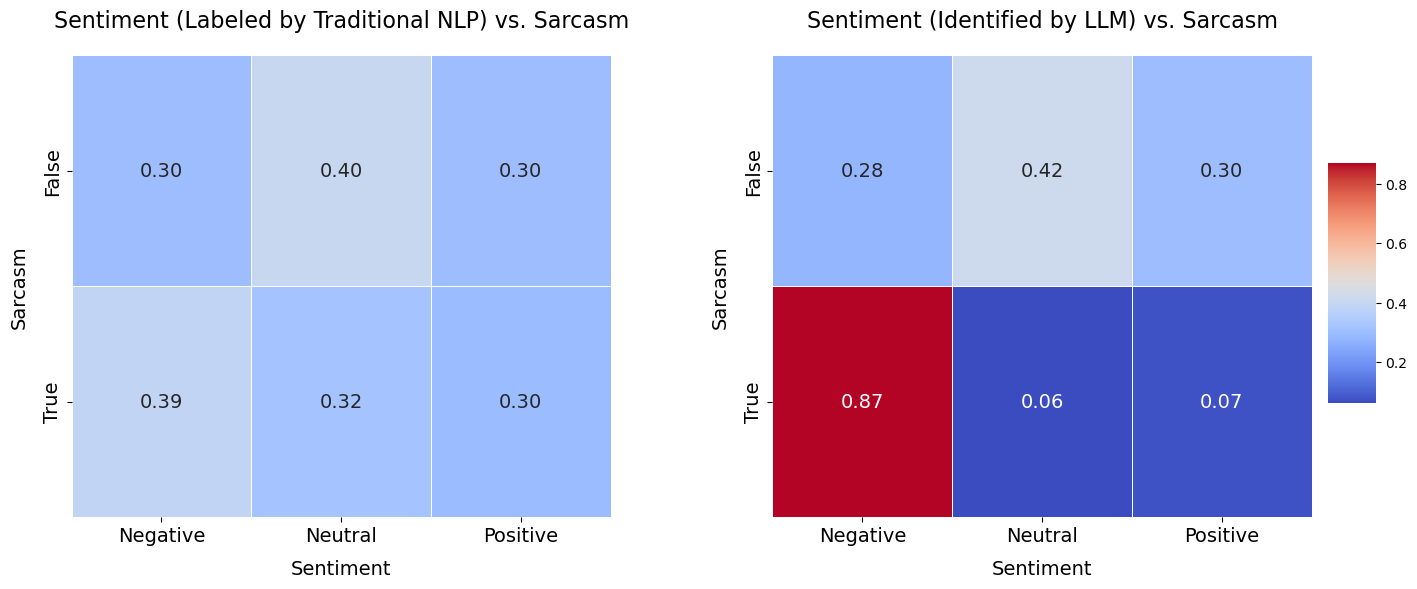}
    \caption{Comparison of Sentiment and Sarcasm Labels: Traditional NLP (left) vs. LLM (right)}
    \label{fig:sentiment_sarcasm_labels}
\end{figure}

\begin{figure}[pos=!htbp]
    \centering
    \includegraphics[width=\textwidth]{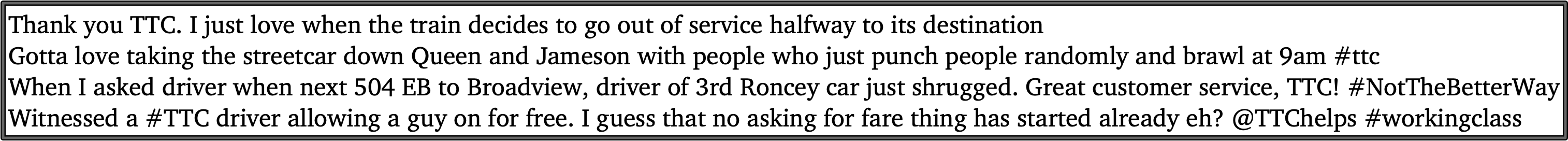}
    \caption{Examples where traditional NLP methods identified posts as positive, but our proposed methods identified them as negative and sarcastic.}
    \label{fig:sentiment_sarcasm}
\end{figure}

Figure~\ref{fig:sentiment_sarcasm_labels} compares sentiment and sarcasm labels extracted by traditional NLP methods and LLM. The left heatmap, using traditional NLP, shows an implausible balance between sentiment and sarcasm labels. Conversely, the LLM-labeled heatmap reveals that most sarcastic tweets are labeled with negative sentiment, demonstrating the LLM's advanced capability in detecting and interpreting sarcasm, as well as sentiment identification.
Figure~\ref{fig:sentiment_sarcasm} illustrates examples where traditional NLP methods identified the posts as positive. However, our proposed methods correctly identified them as negative and sarcastic.

\begin{figure}[pos=!htbp]
    \centering
    \includegraphics[width=\textwidth]{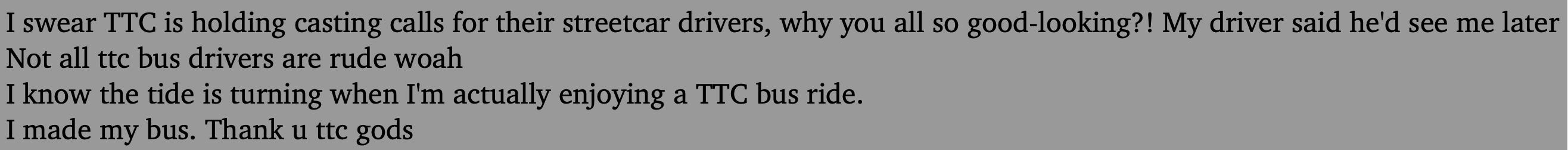}
    \caption{Examples of challenging sentiment analysis where the LLM identified posts as sarcastic but positive.}
    \label{fig:tricky_sentiment}
\end{figure}

Figure~\ref{fig:tricky_sentiment} shows examples of the most challenging cases where the language model identified posts as sarcastic but still positive. These examples demonstrate that even for humans, it is difficult to discern the writer's true opinion without additional context.

\begin{figure}[pos=htbp!]
    \centering
    \includegraphics[width=\linewidth]{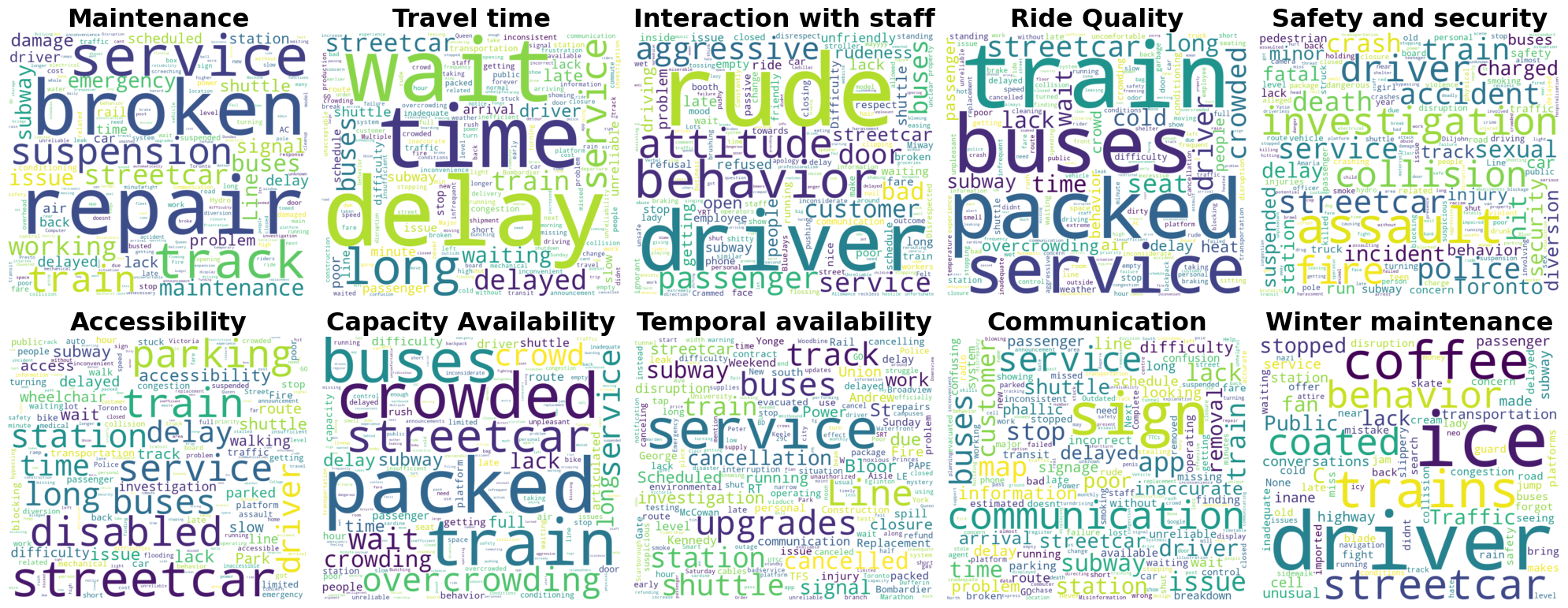}
    \caption{Word Clouds for Different Problem Categories Summarized by LLM}
    \label{fig:wordclouds}
\end{figure}

For system problem extraction, the LLM summarizes the issues mentioned in tweets. Figure~\ref{fig:wordclouds} shows the most frequently mentioned keywords in the LLM-generated problem summaries for each problem category. The alignment of these keywords with the problem categories illustrates the LLM's ability to accurately understand and summarize context.

\begin{figure}[pos=htbp!]
    \centering
    \begin{subfigure}{0.49\textwidth}
        \centering
        \includegraphics[width=\textwidth]{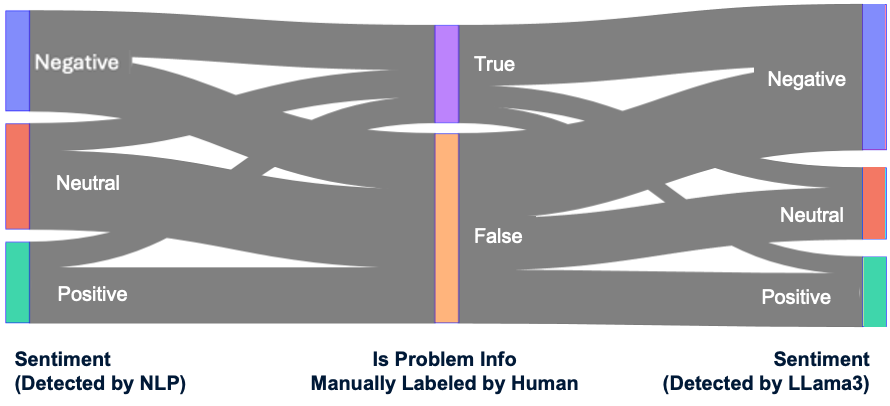}
        \caption{Sentiment vs. Problem Info (Labeled by Human)}
        \label{fig:nlp_human}
    \end{subfigure}
    \hfill
    \begin{subfigure}{0.49\textwidth}
        \centering
        \includegraphics[width=\textwidth]{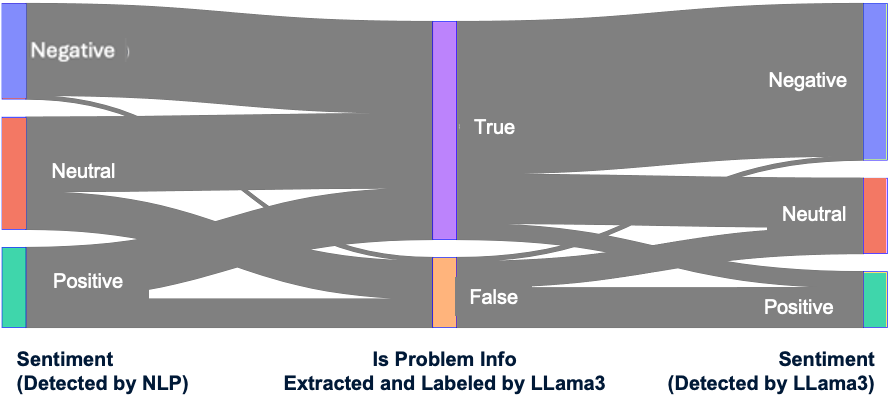}
        \caption{Sentiment vs. Problem Info (Labeled by LLM)}
        \label{fig:nlp_llama3}
    \end{subfigure}
    \caption{Comparison of Sentiment Labels with
    Problem Information Extraction}
    \label{fig:comparison}
\end{figure}

Despite the dataset being (semi-)manually labeled, it contains errors due to the manual labeling process. Figure~\ref{fig:comparison} compares human-labeled and LLM-extracted problem information. The number of records with problems identified by humans is significantly lower than those identified by the LLM, highlighting the limitations of manual labeling.

As shown in Fig.~\ref{fig:nlp_human}, negative tweets, whether estimated by traditional NLP or LLM, are almost evenly found in the 'No problem detected' and 'Problem detected' categories when those categories are labeled by humans, revealing inconsistencies. In contrast, Fig.~\ref{fig:nlp_llama3} shows that using the LLM method for labelling the categories, negative tweets are more accurately identified, aligning better with common sense and behavioral logic.

\subsubsection{Case Study: System Monitoring with LLM Powered Information Extraction Pipeline}

This section demonstrates how the LLM-powered information extraction pipeline can help transit agencies respond promptly and efficiently to system issues.

\begin{figure}[pos=htbp!]
    \centering
    \includegraphics[width=\linewidth]{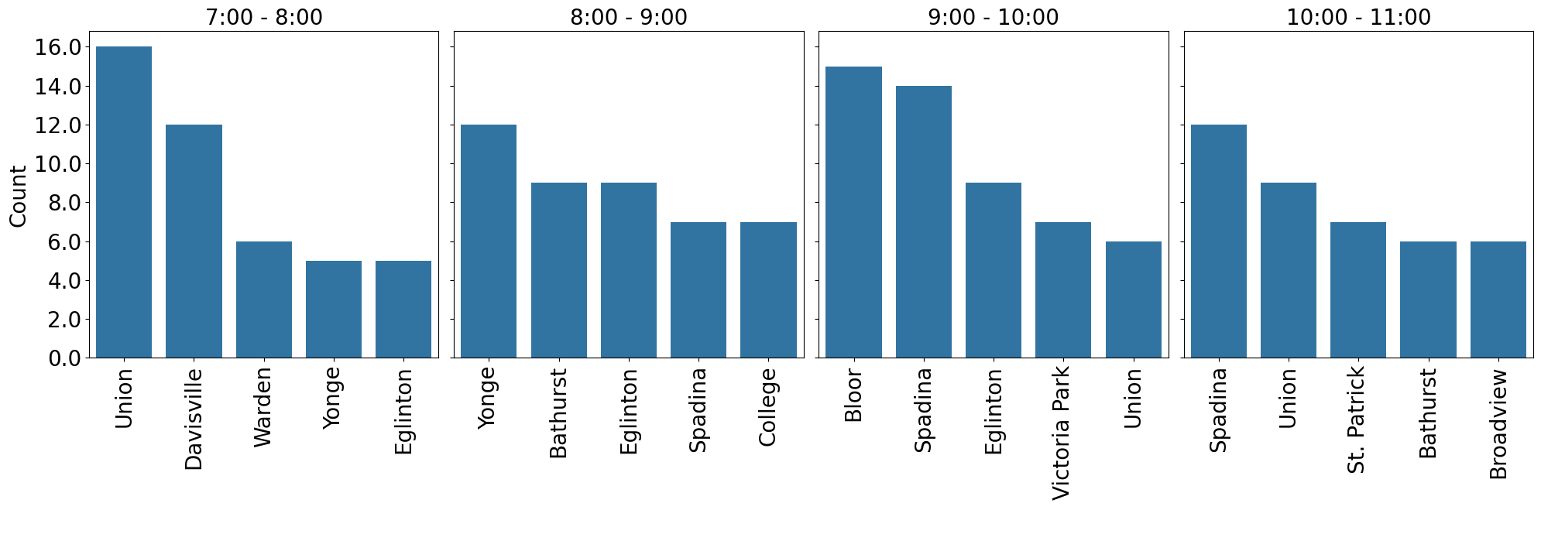}
    \caption{Tweet counts for the five most-mentioned stations from 7 AM to 11 AM}
    \label{fig:tweet_counts}
\end{figure}

Starting with station information, geo-specific data allows for station-level performance analysis. Figure~\ref{fig:tweet_counts} shows tweet counts for the five most-mentioned stations from 7 AM to 11 AM. Union, Yonge, and Eglinton stations are frequently mentioned, with Bloor station peaking from 9 AM to 10 AM, indicating a possible incident.

\begin{figure}[pos=htbp!]
    \centering
    \includegraphics[width=\linewidth]{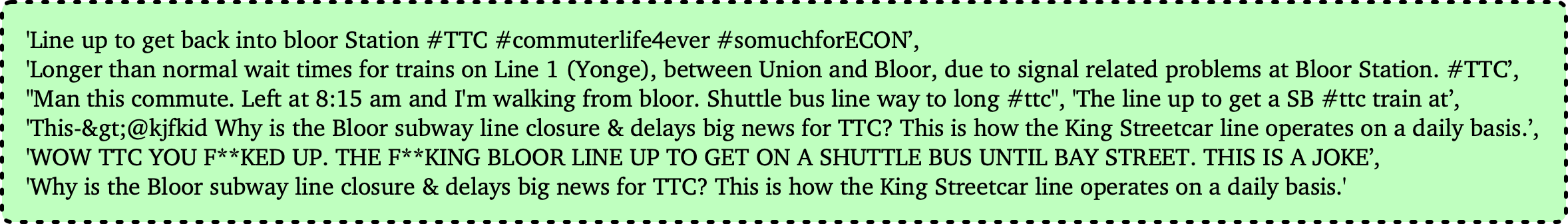}
    \caption{Negative tweets related to Bloor station from 9 to 10 AM}
    \label{fig:bloor_negative_tweets}
\end{figure}

\begin{figure}[pos=htbp!]
    \centering
    \begin{subfigure}[b]{0.45\textwidth}
        \centering
        \includegraphics[width=\textwidth]{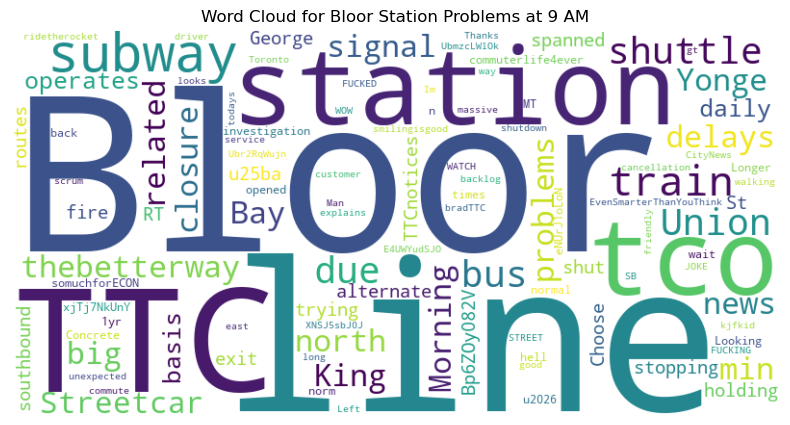}
        \caption{Keywords from tweets related to Bloor station problems at 9 AM}
        \label{fig:direct_tweets}
    \end{subfigure}
    \hfill
    \begin{subfigure}[b]{0.45\textwidth}
        \centering
        \includegraphics[width=\textwidth]{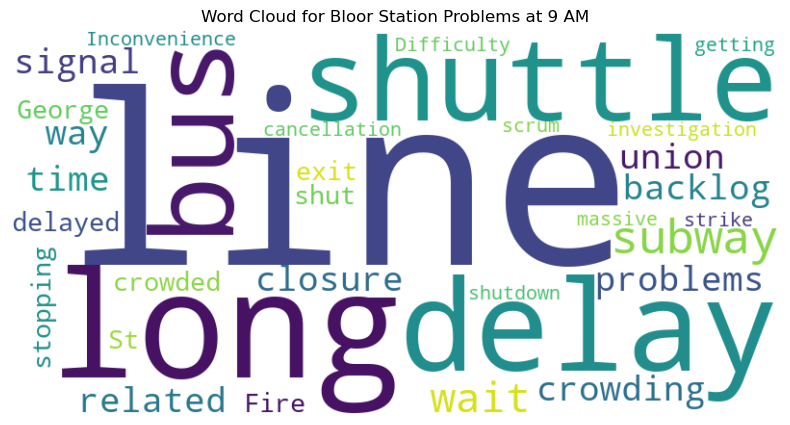}
        \caption{Keywords from LLM summarization of Bloor station problems at 9 AM}
        \label{fig:llm_summary}
    \end{subfigure}
    \caption{Comparison of keywords related to Bloor station problems at 9 AM extracted directly from tweets and from LLM summarization}
    \label{fig:wordclouds_comparison}
\end{figure}

Examining negative tweets related to Bloor station between 9 AM and 10 AM (Figure~\ref{fig:bloor_negative_tweets}), we identify significant issues. Figure~\ref{fig:wordclouds_comparison} compares keywords extracted directly from tweets and LLM summaries. Traditional NLP methods fall short in clarity, whereas LLM summarization clearly identifies the main issue: unusually long lines for shuttle buses, indicating a capacity problem during the morning peak hours. With this information, transit authorities can further investigate using onsite cameras and implement better solutions to address the passenger overflow. 

This example shows how our advanced techniques provide actionable insights for real-time problem-solving.

\section{Conclusion and Future Work}

The LLM-powered information extraction pipeline has demonstrated significant advantages in analyzing social media data for transit systems. This pipeline enhances sentiment analysis by simultaneously detecting sentiment and sarcasm, providing a more accurate understanding of public opinions.

Additionally, the pipeline facilitates actionable insights for transit agencies by transitioning from system-level to station-level analysis. By extracting station-specific information and summarizing potential issues mentioned in social media posts, it enables targeted interventions and improvements.

Moreover, using this LLM-based information extraction pipeline reduces the dependency on pre-identified labels in datasets. By adjusting the prompts and incorporating relevant external guidance documents, the pipeline can extract more useful information with less human effort, broadening the scope of analysis.

However, there are limitations and areas for future research. The TTC-related tweets dataset used in this study lacks comprehensive human annotations for sentiment classification and sarcasm detection. Additionally, station information was not pre-extracted, and there was no human review of labeling results. Building a more robust dataset with high-quality annotations is crucial. This would provide ground truth for classification and extraction tasks and benefit future LLM studies in the transit domain by offering a reliable dataset for training and fine-tuning models, serving as a benchmark for LLM applications in public transit.

Furthermore, refining the prompts used in the LLM is necessary to ensure a more consistent output format, which would enhance the information aggregation process and minimize the loss of relevant information.

Lastly, the LLM used for information extraction, Llama 3, is a high-performance model with 8 billion parameters, requiring substantial computing resources. Future work will focus on employing techniques such as Knowledge Distillation to transfer knowledge from larger models to smaller ones, thereby improving performance and efficiency for handling a relatively limited range of tasks.
\bibliographystyle{cas-model2-names}
\bibliography{conference_0418_2}

\end{document}